\documentclass[10pt,twocolumn,letterpaper]{article}

\usepackage{cvpr}             
\usepackage{graphicx}
\usepackage{amsmath}
\usepackage{amssymb}
\usepackage{booktabs}
\usepackage{mathtools, cuted}
\usepackage{mathrsfs}
\usepackage{breqn}

\usepackage{array}
\usepackage{amsfonts}
\usepackage[margin=1.0in]{geometry}
\usepackage{tabularx}
\usepackage{multirow}
\usepackage{url}
\usepackage{makecell}
\newcolumntype{P}[1]{>{\centering\arraybackslash}p{#1}}
\usepackage{caption}
\usepackage{subcaption}
\usepackage[pagebackref,breaklinks,colorlinks]{hyperref}
\usepackage[capitalize]{cleveref}
\crefname{section}{Sec.}{Secs.}
\Crefname{section}{Section}{Sections}
\Crefname{table}{Table}{Tables}
\crefname{table}{Tab.}{Tabs.}

\begin{document}

\title{Deep Axial Hypercomplex Networks}

\author{Nazmul Shahadat, Anthony S.\ Maida\\
University of Louisiana at Lafayette\\
Lafayette LA 70504, USA\\
{ nazmul.ruet@gmail.com, maida@louisiana.edu}}
\maketitle

\begin{abstract}
Over the past decade, deep hypercomplex-inspired networks have enhanced feature extraction for image classification by enabling weight sharing across input channels. Recent works make it possible to improve representational capabilities by using hypercomplex-inspired networks which consume high computational costs. This paper reduces this cost by factorizing a quaternion 2D convolutional module into two consecutive vectormap 1D convolutional modules. Also, we use 5D parameterized hypercomplex multiplication based fully connected layers. Incorporating both yields our proposed hypercomplex network, a novel architecture that can be assembled to construct deep axial-hypercomplex networks (DANs) for image classifications. We conduct experiments on CIFAR benchmarks, SVHN, and Tiny ImageNet datasets and achieve better performance with fewer trainable parameters and FLOPS. Our proposed model achieves almost 2\% higher performance for CIFAR and SVHN datasets, and more than 3\% for the ImageNet-Tiny dataset and takes six times fewer parameters than the real-valued ResNets. Also, it shows state-of-the-art performance on CIFAR benchmarks in hypercomplex space.
\end{abstract}

\section{Introduction}
\label{sec:intro}

Convolutional neural networks (CNNs) and hypercomplex CNNs (HCNNs) for image classification form a hierarchical design where different layers extract different levels of feature representation. CNNs have shown significant success in recent decades \cite{buyssens2012multiscale,javanmardi2021computer}. In vision tasks, these CNN-based feature extraction designs can be improved in regard to working with multi-dimensional data. To enhance the CNNs ability, HCNNs have been used which treat the multi-dimensional data as a cohesive entity by applying cross-channel weight sharing to discover cross-channel relationships \cite{parcollet2018quaternion,parcollet2019quaternion,gaudet2018deep,gaudet2021removing}. Also, implementations in hypercomplex space provide more advantages \cite{arjovsky2016unitary,danihelka2016associative,hirose2012generalization,nitta2002critical}. It has also been shown that the HCNNs could create better output representations \cite{nitta2002critical,shahadat2021adding,shahadat_2021}. 

Recently, HCNNs with various dimensions like 2D HCNNs \cite{xin2020complex}, 4D HCNNs \cite{gaudet2018deep,parcollet2018quaternion,parcollet2019quaternion}, 8D HCNNs \cite{wu2020deep}, or generalized HCNNs \cite{gaudet2021removing}, have been studied and have hypercomplex properties. The reason behind the success of HCNNs is that they capture the cross-channel relationships \cite{parcollet2018quaternion,parcollet2019quaternion,gaudet2018deep,gaudet2021removing,shahadat2021adding}. Among them, quaternion networks have a set of algebra operations and they have outperformed than the other HCNNs. Stacking quaternion convolutional coherent layers have achieved better representational feature maps and have shown promising results in vision tasks \cite{parcollet2018quaternion,gaudet2018deep,shahadat2021adding}. These networks are cost-effective compared to real-valued CNNs and fully connected networks. But still, they are very expensive for large inputs like vision tasks. 

This work uses an axial hypercomplex network that: 1) handles multidimensional inputs; 2) applies weight sharing across input channels; 3) captures cross-channel correlations; 4) reduces computational costs; and 5) increases validation accuracy performance for image classification datasets. The main idea of this work is to decompose hypercomplex 2D convolutional operation
into two consecutive vectormap 1D convolutional operations
. By splitting 2D spatial convolution operation into height-axis and width-axis spatial convolution, it enables the model to reduce cost once again. Additionally, we apply a quaternion-based stem layer, and parameterized hypercomplex multiplication (PHM) based fully connected layer to get better representation and better generalization performance.

\begin{figure*}
    \centering
    \includegraphics[width=.90\linewidth]{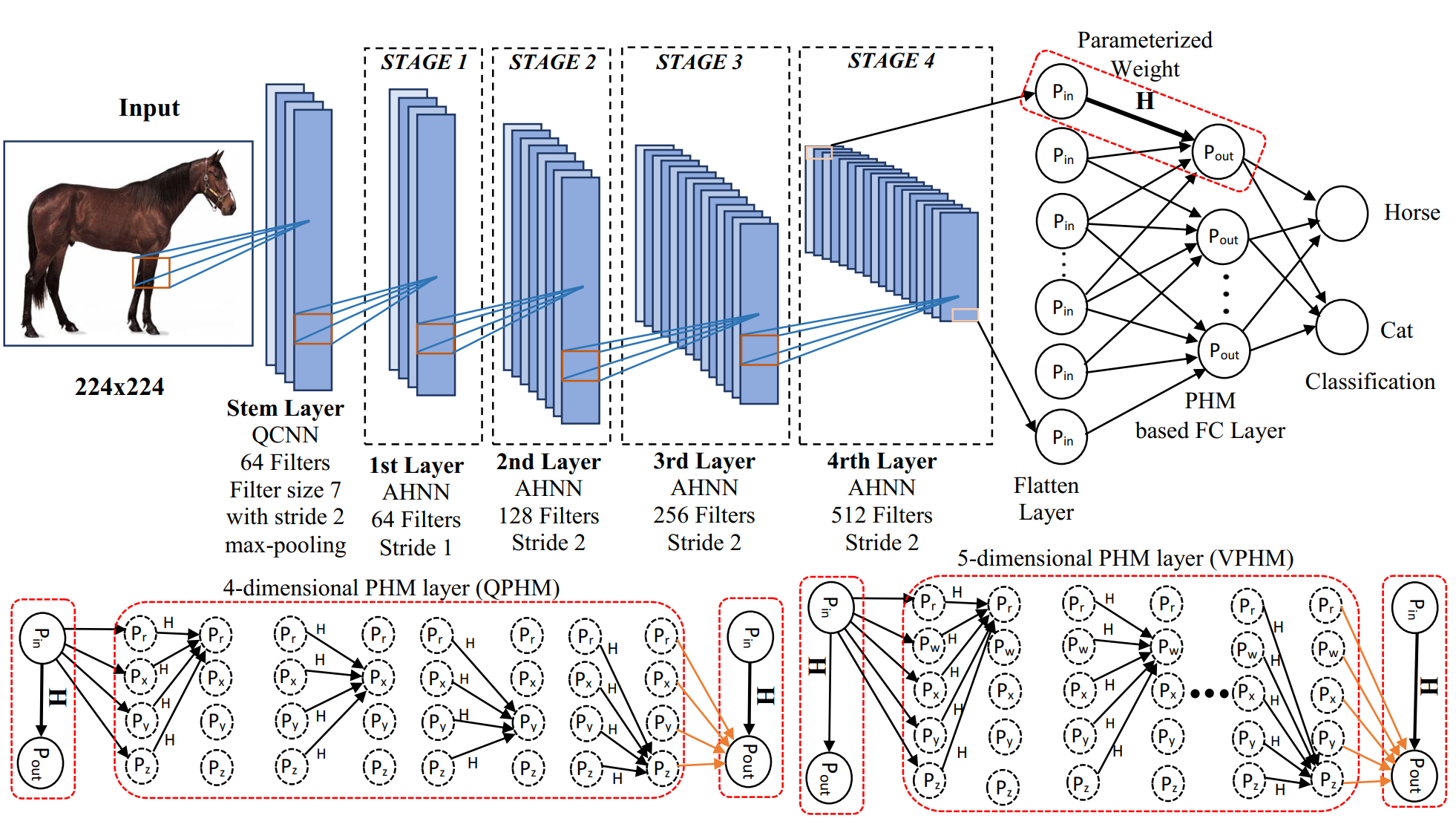}
    \caption{Proposed axial-hypercomplex network with PHM-based fully-connected layer in backend. ``AHNN'' stands for axial-hypercomplex neural network bottleneck block which is described in Figure \ref{fig:AxialHypercomplexBlock}. Here, $Q_{in} = Q_r+Q_w+Q_x+Q_y+Q_z$, $H = H_r+H_w+H_x+H_y+H_z$, and $Q_{out} = Q_{ro}+Q_{wo}+Q_{xo}+Q_{yo}+Q_{zo}$ are the input, hypercomplex parameterized weight, and output, respectively. For the calculation of $H$ see the ``PHM Layer'' section.}
    \label{fig:AxialHypercomplexNet}
\end{figure*}

This paper conducts extensive experiments that show the effectiveness of our novel axial hypercomplex networks on four image classification datasets. 
Our novel contribution is a new model that factorizes the two-dimensional spatial hypercomplex convolutional operation into two one-dimensional operations along the height-axis and width-axis sequentially. 
Our contributions are:
\begin{itemize}
    \item Replacing the spatial $3\times3$ QCNN in the bottleneck block of quaternion ResNets using two VCNNs 
    and showing the effectiveness of the proposed networks.
    \item Applying QCNN in the stem layer (the first layer of the network), resulting in a quaternion-stem model.
    \item Like QPHM \cite{shahadat_2021}, applying PHM-based dense layer in the backend of the network.
\end{itemize} 
This proposed axial hypercomplex ResNets outperformed the baseline networks for classification datasets which is shown in Tables \ref{tab_resultCifar}, \ref{tab_resultSVHN}, and \ref{tab_resultImageNet}. Our experiments show that the proposed model achieves state-of-the-art results with far fewer trainable parameters, and FLOPS for CIFAR benchmarks in hypercomplex space.
\begin{figure*}
    \centering
    \includegraphics[width=0.75\textwidth,height=170pt]{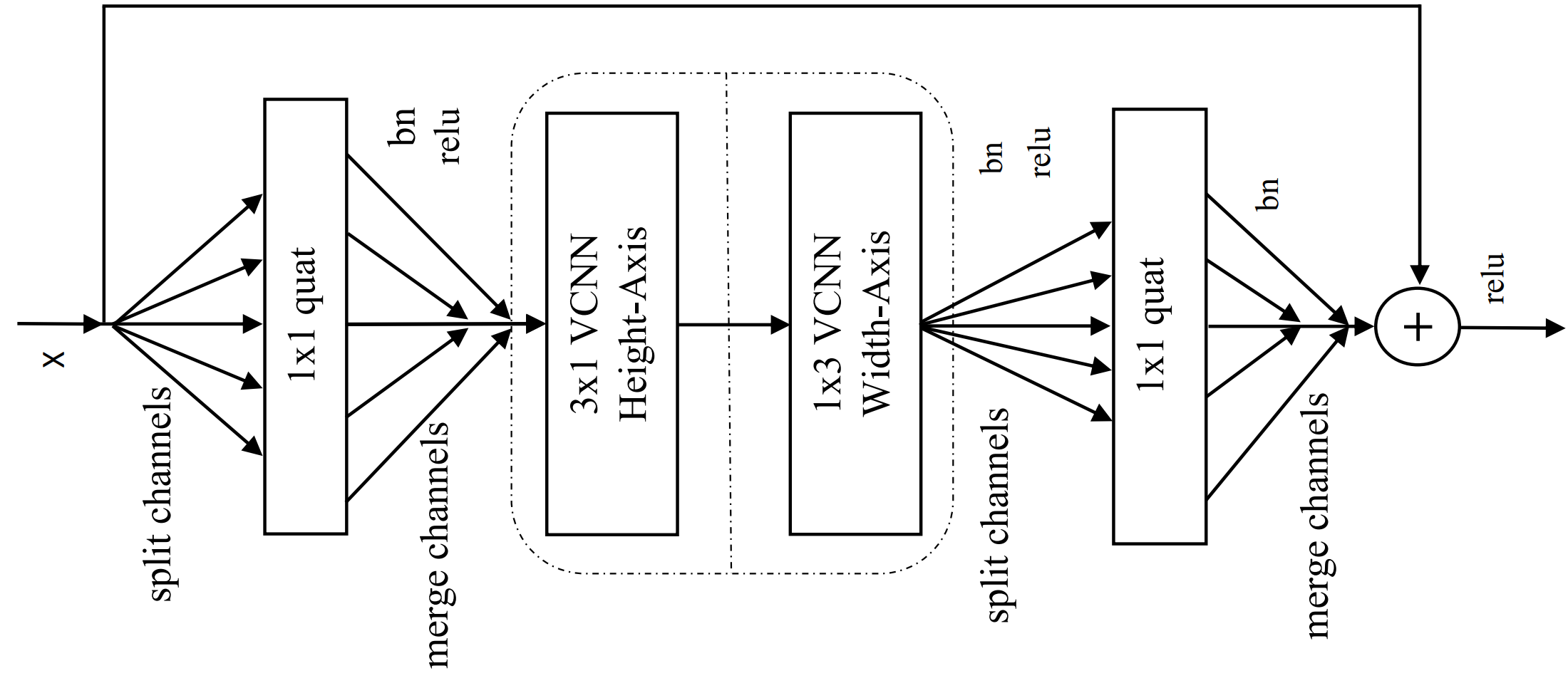}
    \caption{AHNN bottleneck block used in our proposed axial-hypercomplex networks. ``bn", ``quat'', and ``VCNN" stand for batch normalization, quaternion CNN, and vectormap CNN, respectively.}
    \label{fig:AxialHypercomplexBlock}
\end{figure*}

\section{Background and Related Work}
\subsection{Quaternion Convolution}
\label{subsection_quatconv}
The deep quaternion CNN extends of complex CNNs \cite{trabelsi2017deep}. This section explains cross channel weight sharing. \cite{gaudet2018deep} and \cite{parcollet2018quaternion} extended the principles of quaternion  convolution operations, and weight initialization. Quaternion number system is formed as, $ 
Q =  r + \textit{i}x + \textit{j}y + \textit{k}z~;~r,x,y,z \in \mathbb{R} 
$
where, $r$, $x$, $y$, and $z$ are real values and $i, j,$ and $k$ are imaginary.
Quaternion convolution between quaternion filter matrix $F$
and quaternion input vector $M$
, is defined as \cite{gaudet2018deep}:
\begin{equation}
    \begin{aligned}
        M \circledast F = (\mathbf{O_r}, \mathbf{O_i}, \mathbf{O_j}, \mathbf{O_k})\\
        = (\mathbf{M_r}*\mathbf{F_r} - \mathbf{M_i}*\mathbf{F_i} - \mathbf{M_j}*\mathbf{F_j} -\mathbf{M_k}*\mathbf{F_k},\\
        \mathbf{M_i}*\mathbf{F_r} + \mathbf{M_r}*\mathbf{F_i} + \mathbf{M_j}*\mathbf{F_k} - \mathbf{M_k}*\mathbf{F_j},\\
        \mathbf{M_j}*\mathbf{F_r} + \mathbf{M_r}*\mathbf{F_j} + \mathbf{M_k}*\mathbf{F_i} - \mathbf{M_i}*\mathbf{F_k},\\ 
        \mathbf{M_k}*\mathbf{F_r} + \mathbf{M_r}*\mathbf{F_k} + \mathbf{M_i}*\mathbf{F_j} - \mathbf{M_j}*\mathbf{F_i})
        \label{equ:3quatMultiplication}
    \end{aligned}
\end{equation}
\noindent
where, $\mathbf{M \circledast F}$, and all others are quaternion numbers. $\mathbf{O_r}$ is the real part, and $\mathbf{O_i}$, $\mathbf{O_j}$, and $\mathbf{O_k}$ are the imaginary parts. Although there are 16 real-valued convolutions in Equation \ref{equ:3quatMultiplication}, there are only four kernels that are reused. The weight sharing happens this way \cite{parcollet2019quaternion} which forces the model to learn cross-channel interrelationships. According to the quaternion definition, a quaternion layer can accept  four or $m$ numbers of input channels, where $m$ is divisible by four. To process $m$ input channels ($m\geq4$), $m/4$ number of independent quaternion convolution modules is required. Also, there are $m/4$ weight sets where each module has its own weight sets. Cross-channel weight sharing allows discovering of cross-channel input correlations. Our weight initialization was the same as \cite{gaudet2018deep}.

\subsection{Vectormap Convolution}
\label{subsection_vectconv}
We explain 3D generalized hypercomplex networks or VCNNs as 
VCNNs are used in our proposed models. The VCNN is more flexible as it doesn't require 4D. However, still using cross channel weight sharing this is seen in $3 \times 3$ matrix used in Equation \ref{e:vectconv}, only three filters A, B, and C are used. The Vectormap convolution operation is defined as:
\begin{equation}
\begin{bmatrix}
 \mathcal{R}(\textbf{$M$}\ast \textbf{$F$}) \\ 
 \mathcal{I}(\textbf{$M$}\ast \textbf{$F$}) \\ 
 \mathcal{J}(\textbf{$M$}\ast \textbf{$F$}) 
\end{bmatrix}
= L \odot
\begin{bmatrix}
 \textbf{A} & \textbf{B} & \textbf{C} \\
 \textbf{C} & \textbf{A} & \textbf{B} \\
 \textbf{B} & \textbf{C} & \textbf{A} 
\end{bmatrix}
\ast
\begin{bmatrix}
 \textbf{x} \\
 \textbf{y} \\
 \textbf{z} 
\end{bmatrix}
\label{e:vectconv}
\end{equation}
where, $\mathbf{A}$, $\mathbf{B}$, and $\mathbf{C}$ are real-valued kernels, $\mathbf{x}$, $\mathbf{y}$, and $\mathbf{z}$ being real-valued vectors, and L is a learnable matrix, $L \in \mathbb{R}^{D_3 \times D_3}$; where $D_3$ stands for 3-dimensional input channels. The initial value of this matrix L is defined as:
\begin{equation}
L = 
\begin{bmatrix}
 \textbf{1} & \textbf{1} & \textbf{1} \\
 \textbf{-1} & \textbf{1} & \textbf{1} \\
 \textbf{-1} & \textbf{1} & \textbf{1} 
\end{bmatrix}
\label{e:vectLvalue}
\end{equation}
Our weight initialization follows \cite{gaudet2021removing}.

\subsection{PHM Layer}
\label{subsection_PHM}
Parameterized hypercomplex multiplication is another form of generalized hypercomplex network, explained in \cite{zhang2021beyond}. As we use this PHM layer only in the fully connected (FC) layer, our explanation is restricted to this PHM-based dense layer. It is defined as, 
$y = Hx + b$, where $H \in  \mathbb{R}^{k \times d}$ represents the PHM layer and it is calculated as, $H = \sum_{i=1}^n \mathbf{I}_i \otimes \mathbf{A}_i$, where $\mathbf{I}_i\in\mathbb{R}^{n\times n}$
and $\mathbf{A}_i \in \mathbb{R}^{k/n\times d/n}$ are learnable parameter matrices and $i = 1\ldots n$ ($n=4$ or $5$). These matrices can be reused which leads to parameter reduction. Also, the $\otimes$ represents the Kronecker product. The flattened layer which is the output of the CNN network is used as an input to the PHM FC layer. These inputs are split as, $Q_{in} = Q_r + Q_w + Q_x + Q_y + Q_z$ and the outputs are merged into $Q_{out}$ as, $
 Q_{out} = Q_{ro} + Q_{wo} + Q_{xo} + Q_{yo} + Q_{zo}$ for 5D hypercomplex. The 4D hypercomplex parameter matrix is discussed in \cite{zhang2021beyond} which expresses the Hamiltonian product, and the 5D hypercomplex parameter matrix of PHM operation is explained in \cite{shahadat_2021}. This 5D parameter matrix is used to construct a 5D PHM FC layer which preserves all properties of the PHM layer and hypercomplex networks. This work uses 5D PHM layer.

\section{Proposed Axial Hypercomplex Networks}
\label{sec:AxialHypercomplexArchi} 
Complex convolutional neural networks (CCNNs), QCNNs, Octonions convolutional neural networks (OCNNs), VCNNs, and PHM are the versions of HCNNs that provide all advantages of HCNNs like weight sharing across input channels, and the ability to discover cross channel correlations. These HCNNs perform better with fewer trainable parameters for vision applications. But, they are still computationally expensive. For vision tasks, these HCNNs take $\mathcal{O}(N^2)$ resources for an image of length $N$ where $N$ is the flattened pixel set. For a 2D image of height $h$ and width $w$,  where $N=hw$, and $h=w$, the computational cost is $\mathcal{O}((hw)^2) = \mathcal{O}(h^2 w^2) = \mathcal{O}(h^4)$. 

\begin{table*}
\centering
\begin{tabular}{c|c|c|c|c|c} 
\hline
Layer & \shortstack{Output\\size} &  \shortstack{Deep Quaternion\\ResNet} & \shortstack{Vectormap\\ResNet} & \shortstack{QPHM} & \shortstack{Axial\\Hypercomplex} \\ \hline
\shortstack{Stem} & 32x32 & 3x3Q, 120, std=1 & 3x3V, 120, std=1 & 3x3Q, 120, std=1 & 3x3Q, 120, std=1  \\ \hline

\shortstack{\vspace{2pt} Bottleneck\\ group 1} &$32\mathrm{x}32$& 
 $ \begin{bmatrix} 1\mathrm{x}1\mathrm{Q},120 \\ 3\mathrm{x}3\mathrm{Q},120 \\ 1\mathrm{x}1\mathrm{Q},480 \end{bmatrix}$\hspace{-3pt}${\times} 3 $ & 
$ \begin{bmatrix} 1\mathrm{x}1\mathrm{V}, 120 \\ 3\mathrm{x}3\mathrm{V}, 120 \\ 1\mathrm{x}1\mathrm{V}, 480 \end{bmatrix}$\hspace{-3pt}${\times} 3 $ & 
$\begin{bmatrix} 1\mathrm{x}1\mathrm{QP},120 \\ 3\mathrm{x}3\mathrm{QP},120 \\ 1\mathrm{x}1\mathrm{QP},480 \end{bmatrix}$\hspace{-3pt}${\times} 3 $ &
$ \begin{bmatrix} 1\mathrm{x}1\mathrm{Q},120 \\ 3\mathrm{x}1\mathrm{AV},120 \\
1\mathrm{x}3\mathrm{AV},120 \\
1\mathrm{x}1\mathrm{Q},480 \end{bmatrix}$\hspace{-3pt}${\times} 3 $ 
\\ \hline

\shortstack{\vspace{2pt} Bottleneck\\group 2} & $16\mathrm{x}16$ &  
$\begin{bmatrix} 1\mathrm{x}1\mathrm{Q}, 240 \\ 3\mathrm{x}3\mathrm{Q}, 240 \\ 1\mathrm{x}1\mathrm{Q}, 960 \end{bmatrix}$\hspace{-3pt}${\times} 4 $ &
$ \begin{bmatrix} 1\mathrm{x}1\mathrm{V}, 240 \\ 3\mathrm{x}3\mathrm{V}, 240 \\ 1\mathrm{x}1\mathrm{V}, 960 \end{bmatrix}$\hspace{-3pt}${\times} 4 $ &
$\begin{bmatrix} 1\mathrm{x}1\mathrm{QP}, 240 \\ 3\mathrm{x}3\mathrm{QP}, 240 \\ 1\mathrm{x}1\mathrm{QP}, 960 \end{bmatrix}$\hspace{-3pt}${\times} 4 $ &
$ \begin{bmatrix} 1\mathrm{x}1\mathrm{Q},240 \\ 3\mathrm{x}1\mathrm{AV},240 \\
1\mathrm{x}3\mathrm{AV},240 \\
1\mathrm{x}1\mathrm{Q},960 \end{bmatrix}$\hspace{-3pt}${\times} 4$ \\ \hline

\shortstack{\vspace{2pt} Bottleneck\\group 3} & $8\mathrm{x}8$ &
$\begin{bmatrix} 1\mathrm{x}1\mathrm{Q}, 480 \\ 3\mathrm{x}3\mathrm{Q}, 480 \\ 1\mathrm{x}1\mathrm{Q}, 1920 \end{bmatrix}$\hspace{-3pt}${\times} 6 $& 
$ \begin{bmatrix} 1\mathrm{x}1\mathrm{V}, 480 \\ 3\mathrm{x}3\mathrm{V}, 480 \\ 1\mathrm{x}1\mathrm{V}, 1920 \end{bmatrix}$\hspace{-3pt}${\times} 6 $& 
$\begin{bmatrix} 1\mathrm{x}1\mathrm{QP}, 480 \\ 3\mathrm{x}3\mathrm{QP}, 480 \\ 1\mathrm{x}1\mathrm{QP}, 1920 \end{bmatrix} $\hspace{-3pt}${\times} 6 $ &
$ \begin{bmatrix} 1\mathrm{x}1\mathrm{Q},480 \\ 3\mathrm{x}1\mathrm{AV},480 \\
1\mathrm{x}3\mathrm{AV},480 \\
1\mathrm{x}1\mathrm{Q},1920 \end{bmatrix}$\hspace{-3pt}${\times} 6 $ 
\\ \hline
\shortstack{\vspace{2pt} Bottleneck\\group 4} & $4\mathrm{x}4$ & 
$\begin{bmatrix} 1\mathrm{x}1\mathrm{Q}, 960 \\ 3\mathrm{x}3\mathrm{Q},960\\ 1\mathrm{x}1\mathrm{Q},3840\end{bmatrix}$\hspace{-3pt}${\times}3 $ &
$ \begin{bmatrix} 1\mathrm{x}1\mathrm{V}, 960 \\ 3\mathrm{x}3\mathrm{V}, 960 \\ 1\mathrm{x}1\mathrm{V}, 3840 \end{bmatrix}$\hspace{-3pt}${\times} 3 $ &
$\begin{bmatrix} 1\mathrm{x}1\mathrm{QP}, 960 \\ 3\mathrm{x}3\mathrm{QP}, 960 \\ 1\mathrm{x}1\mathrm{QP}, 3840 \end{bmatrix} $\hspace{-3pt}${\times} 3 $ &
$ \begin{bmatrix} 1\mathrm{x}1\mathrm{Q},960 \\ 3\mathrm{x}1\mathrm{AV},960 \\
1\mathrm{x}3\mathrm{AV},960 \\ 1\mathrm{x}1\mathrm{Q},3840 \end{bmatrix}$\hspace{-3pt}${\times}3 $ \\ \hline

\shortstack{Pooling layer}& $1\mathrm{x}1\mathrm{x}100$ & \multicolumn{3}{c}{global average-pool, 100 outputs}& \\ \hline

Output& $1\mathrm{x}1\mathrm{x}100$ & \multicolumn{2}{c|}{fully connected layer, softmax}& \multicolumn{2}{c}{5D PHM layer} \\ \hline
\end{tabular}
\caption{The 50-layer architectures tested on CIFAR-100: quaternion ResNet \cite{gaudet2018deep,gaudet2021removing}, 
vectormap ResNet \cite{gaudet2021removing}, QPHM \cite{shahadat_2021}, and our proposed axial-hypercomplex networks. Input is a 32x32x3 color image for CIFAR benchmarks. The number of stacked bottleneck modules is specified by multipliers. ``Q'', ``V'', ``QP'', ``AV'' and ``std'' denote quaternion convolution, 3D vectormap convolution, QPHM (quaternion networks with 4D PHM layer), axial vectormap convolution, and stride correspondingly. Integers (e.g., 120, 240) denote the number of output channels. PHM layer stands for parameterized hypercomplex multiplication layer. This work uses 5D PHM based FC layer.
\label{tab_archiTable50}
}
\end{table*}
This section describes our proposed axial-hypercomplex model in Figures~\ref{fig:AxialHypercomplexNet} and \ref{fig:AxialHypercomplexBlock} to reduce the computational cost. Axial networks were first used in \cite{ho2019axial,wang2020axial}. To implement our proposed model, we followed the assumption that images are approximately square where the pixel count of $h$ and $w$ are the same, and both are much less than the pixel count of $hw$ \cite{wang2020axial}. To translate a quaternion convolutional bottleneck block to an axial-hypercomplex bottleneck block, we replace the $3\times3$ spatial quaternion convolutional operation by two axial vectormap convolutional neural network (VCNN) layers. These layers are applied to the height axis (3 channels 3x1 VCNN layer) and width axis (3 channels 1x3 VCNN layer) sequentially. The two $1\times1$ quaternion convolutional layers remain unchanged like the original QCNNs \cite{gaudet2018deep}. The $1\times 1$ QCNNs are responsible to reduce and then increase the number of channels. This forms our proposed axial-hypercomplex bottleneck block seen in Figure~\ref{fig:AxialHypercomplexBlock}. This block is stacked multiple times to construct axial-hypercomplex ResNets. 

Axial-hypercomplex models only work on one dimension at a time but the input images are 2-dimensional. For two-dimensional vision tasks, a square 2D input where $h=w$, so $w^2 = N$, where $N$ is the sequence length of the flattened pixel set, is split into two 1D vectors. The 3-channel VCNN operation is first applied along the 1D input image region of length $h$ and then applied along the 1D input image region of length $w$. These two 1D operations finally merged together reduces cost to $\mathcal{O}(h \cdot  
h^2)=\mathcal{O}(h^3)$ from the HCNNs cost of $\mathcal{O}(h^4)$.

Each quaternion convolution accepts four channels of input and produces four channels of output. Hence, the required number of $1\times1$ quaternion conv2d modules equals the number of input channels divided by four.  The set of output channels of down-sampled $1\times 1$ quaternion is merged into input to the axial VCNN modules, and the output channels of axial VCNN modules are split into groups of four again for $1\times1$ up-sampled quaternion conv2d layer \cite{gaudet2018deep,shahadat2021adding}. One quaternion 2D convolution is applied to each group of four channels and one vectormap 2D convolution is applied to each group of three channels. Like vectormap, each axial vectormap module takes three input channels. Thus, the weight-sharing is compartmentalized into groups of four input channels and then groups of three input channels.

For better representation, a quaternion convolution layer is also used in the stem layer (first layer of the network) as a quaternion-based frontend layer and the fully-connected dense layer as a PHM-based backend layer of deep axial-hypercomplex networks (DANs). Figure~\ref{fig:AxialHypercomplexNet} illustrates our proposed axial-hypercomplex network architecture.

\begin{table*}
\centering
\begin{tabular}{|l|c|c|c|c|c|c|} \hline
Model Name & Layers & Dataset & Params & FLOPS & Latency & \shortstack{Validation\\ Accuracy} \\ \hline

ResNet \cite{he2016deep} & & & 40.9M & 2.56G & 0.86ms & 94.68 \\
ResNet-with-QPHM \cite{shahadat_2021} &&  & 40.8M & 2.55G & 0.64ms & 95.32 \\
Quaternion \cite{gaudet2018deep} &&  & 10.2M & 1.11G & 0.65ms & 94.89 \\
Vectormap \cite{gaudet2021removing}&26& CIFAR10  & 13.6M & 1.09G & 0.65ms & 94.76\\
QPHM \cite{shahadat_2021}  & & & 10.2M & 1.10G & 0.64ms & 95.26 \\ 
VPHM \cite{shahadat_2021}  & & & 13.6M & 1.08G & 0.67ms & 95.15 \\
\textbf{Axial-Hypercomplex}  & & & \textbf{6.2M} & \textbf{1.06G} & \textbf{0.68ms} & \textbf{95.91-95.85}\\\hline

ResNet \cite{he2016deep} & & & 57.8M & 3.31G & 1.08ms & 94.95 \\
ResNet-with-QPHM \cite{shahadat_2021} & & & 57.7M & 3.31G & 0.81ms & 95.80 \\
Quaternion \cite{gaudet2018deep} & & & 14.5M & 1.47G & 0.82ms & 95.33 \\
Vectormap \cite{gaudet2021removing} & 35 & CIFAR10 & 19.3M & 1.45G & 0.84ms & 95.06 \\
QPHM \cite{shahadat_2021}  & & & 14.5M & 1.46G & 0.79ms & 95.55 \\
VPHM \cite{shahadat_2021}  & & & 19.3M & 1.44G & 0.82ms & 95.60\\
\textbf{Axial-Hypercomplex}  & & & \textbf{9.2M} & \textbf{1.36G} & \textbf{0.84ms} & \textbf{96.49-96.45} \\\hline

ResNet \cite{he2016deep} & & & 82.5M & 4.57G & 1.32ms & 94.08\\
ResNet-with-QPHM \cite{shahadat_2021} & & & 82.5M & 4.57G & 0.81ms & 95.86 \\
Quaternion \cite{gaudet2018deep} & & & 21.09M & 1.93G & 1.06ms & 95.42\\
Vectormap \cite{gaudet2021removing} & 50& CIFAR10 & 27.6M & 1.93G & 1.13ms & 95.37\\
QPHM \cite{shahadat_2021}  & & & 20.7M & 1.92G & 1.06ms & 95.75\\
VPHM \cite{shahadat_2021}  & & & 27.5M & 1.92G & 1.08ms & 95.76\\
\textbf{Axial-Hypercomplex} & & & \textbf{13.6M} & \textbf{1.75G} & \textbf{1.09ms} & \textbf{96.79-96.71}\\\hline \multicolumn{7}{|l|}{\vspace{-5pt}} \\ \hline

ResNet \cite{he2016deep} & & & 41.2M & 2.56G & 0.89ms & 78.21 \\
ResNet-with-QPHM \cite{shahadat_2021} &&  & 40.9M & 2.56G & 0.64ms & 79.14 \\
Quaternion \cite{gaudet2018deep} &&  & 10.6M & 1.15G & 0.64ms & 77.65 \\
Vectormap \cite{gaudet2021removing}&26&CIFAR100 & 13.6M & 1.15G & 0.64ms & 77.65\\
QPHM \cite{shahadat_2021}  & & & 10.3M & 1.11G & 0.65ms & 78.15 \\ 
VPHM \cite{shahadat_2021}  & & & 13.7M & 1.09G & 0.66ms & 78.14 \\
\textbf{Axial-Hypercomplex}  & & & \textbf{6.2M} & \textbf{1.06G} & \textbf{0.69ms} & \textbf{79.42-79.24}\\\hline

ResNet \cite{he2016deep} & & & 58.1M & 3.31G & 1.07ms& 78.72 \\
ResNet-with-QPHM \cite{shahadat_2021} & & & 57.8M & 3.31G & 0.81ms & 79.65 \\
Quaternion \cite{gaudet2018deep} & & & 14.5M & 1.51G & 0.81ms & 78.96 \\
Vectormap \cite{gaudet2021removing} & 35&CIFAR100& 19.3M & 1.48G & 0.84ms & 79.52 \\
QPHM \cite{shahadat_2021}  & & & 14.5M & 1.47G & 0.82ms & 78.46 \\
VPHM \cite{shahadat_2021}  & & & 19.6M & 1.45G & 0.82ms& 79.86\\
\textbf{Axial-Hypercomplex}  & & & \textbf{9.2M} & \textbf{1.36G} & \textbf{0.85ms} & \textbf{79.93-79.63}\\\hline

ResNet \cite{he2016deep} & & & 82.9M & 4.57G & 1.36ms & 78.95\\
ResNet-with-QPHM \cite{shahadat_2021} & & & 82.6M & 4.57G & 1.09ms & 79.89\\
Quaternion \cite{gaudet2018deep} & & & 21.09M & 1.96G & 1.06ms & 79.17 \\
Vectormap \cite{gaudet2021removing} & 50&CIFAR100& 27.6M & 1.93G & 1.13ms & 79.39 \\
QPHM \cite{shahadat_2021}  & & & 20.7M & 1.93G & 1.05ms & 78.22 \\
VPHM \cite{shahadat_2021}  & & & 27.5M & 1.92G & 1.08ms &79.49 \\
\textbf{Axial-Hypercomplex} & & & \textbf{13.6M} & \textbf{1.75G} & \textbf{1.09ms} & \textbf{80.81-80.75}\\\hline
\end{tabular} 
\caption{Image classification performance on the CIFAR benchmarks for 26, 35, and 50-layer architectures. Here, QPHM, and VPHM define the quaternion networks with PHM FC layer,
and vectormap networks with the PHM FC layer, respectively.}
\label{tab_resultCifar}
\end{table*}

\begin{table*}
\centering
\begin{tabular}{|l|c|c|c|c|c|} \hline
Model Name & Layers  & Params & FLOPS & Latency & Validation Accuracy \\ \hline

ResNet \cite{he2016deep} & & 40.9M & 2.56G & 0.82ms & 96.04 \\
ResNet-with-QPHM \cite{shahadat_2021} &  & 40.8M & 2.56G & 0.62ms & 96.64 \\
Quaternion \cite{gaudet2018deep} &  & 10.2M & 1.11G & 0.66ms & 95.88 \\
Vectormap \cite{gaudet2021removing}&26& 13.6M & 1.10G & 0.66ms & 95.93 \\
QPHM \cite{shahadat_2021}  & & 10.2M & 1.10G & 0.62ms & 95.97 \\ 
VPHM \cite{shahadat_2021}  & & 13.6M & 1.08G & 0.64ms & 96.24 \\
\textbf{Axial-Hypercomplex}  & & \textbf{6.2M} & \textbf{1.06G} & \textbf{0.69ms} & \textbf{97.21-97.05} \\\hline

ResNet \cite{he2016deep} & & 57.8M & 3.31G & 0.98ms& 95.74 \\
ResNet-with-QPHM \cite{shahadat_2021} & & 57.7M & 3.31G & 0.79ms & 96.22 \\
Quaternion \cite{gaudet2018deep} & & 14.5M & 1.47G & 0.84ms & 95.95 \\
Vectormap \cite{gaudet2021removing} & 35& 19.5M & 1.45G & 0.84ms & 95.97 \\
QPHM \cite{shahadat_2021}  & & 14.5M & 1.45G & 0.82ms & 95.99 \\
VPHM \cite{shahadat_2021}  & & 19.3M & 1.44G & 0.82ms & 96.34\\
\textbf{Axial-Hypercomplex}  & & \textbf{9.2M} & \textbf{1.36G} & \textbf{0.85ms} & \textbf{97.25-96.90}\\\hline

ResNet \cite{he2016deep} &  & 82.5M & 4.57G & 1.19ms & 95.76\\
ResNet-with-QPHM \cite{shahadat_2021} &  & 82.5M & 4.57G & 1.04ms & 96.78\\
Quaternion \cite{gaudet2018deep} & & 20.7M & 1.94G & 1.04ms & 96.24\\
Vectormap \cite{gaudet2021removing} & 50 & 27.6M & 1.93G & 1.11ms & 96.39\\
QPHM \cite{shahadat_2021}  &  & 20.7M & 1.93G & 1.04ms & 96.46\\
VPHM \cite{shahadat_2021}  &  & 27.5M & 1.92G & 1.09ms & 96.49\\
\textbf{Axial-Hypercomplex} &  & \textbf{13.6M} & \textbf{1.75G} & \textbf{1.11ms} & \textbf{97.47-97.25} \\\hline
\end{tabular} 
\caption{Image classification performance on the SVHN benchmarks for 26, 35, and 50-layer architectures. Here, QPHM, and VPHM 
define the quaternion networks with PHM FC layer,
and vectormap networks with PHM FC layer, 
respectively.}
\label{tab_resultSVHN}
\end{table*}

\section{Experiment}
We conduct an extensive experiment on four classification datasets to analyze the effectiveness of our proposed axial-hypercomplex model. As QCNNs, VCNNs, residual networks (ResNets), QPHM \cite{shahadat_2021}, and VPHM \cite{shahadat_2021} all are performed 2D spatial convolution operation, therefore we compare our proposed axial hypercomplex networks performance with the above-mentioned baseline models. Among them, all models perform Hamiltonian products like our proposed model except ResNets.

\begin{table*}
\centering
\begin{tabular}{|l|c|c|c|c|c|} \hline
Model Name & Layers  & Params & FLOPS & Latency & Validation Accuracy \\ \hline

ResNet \cite{he2016deep} & & 41.6M & 10.2G & 3.06ms & 57.21 \\
ResNet-with-QPHM \cite{shahadat_2021} &  & 41M & 2.56G & 2.31ms & 57.84 \\
Quaternion \cite{gaudet2018deep} &  & 11.02M & 4.54G & 2.48ms & 53.84 \\
Vectormap \cite{gaudet2021removing}&26& 14.4M & 4.56G & 2.88ms & 56.15\\
QPHM \cite{shahadat_2021}  & & 10.4M & 1.11G & 2.31ms & 54.02 \\ 
VPHM \cite{shahadat_2021}  & & 13.8M & 4.44G & 3.27ms & 53.11 \\
\textbf{Axial-Hypercomplex}  & & \textbf{6.3M} & \textbf{1.06G} & \textbf{2.49ms} & \textbf{58.56-58.06} \\\hline

ResNet \cite{he2016deep} & & 58.5M & 13.2G & 3.21ms& 57.80 \\
ResNet-with-QPHM \cite{shahadat_2021} & & 57.9M & 3.31G & 2.85ms & 59 \\
Quaternion \cite{gaudet2018deep} & & 15.2M & 5.98G & 3.52ms & 54.53 \\
Vectormap \cite{gaudet2021removing} & 35& 20.07M & 5.98G & 3.76ms & 55.99 \\
QPHM \cite{shahadat_2021}  & & 14.6M & 1.47G & 2.88ms & 56.42 \\
VPHM \cite{shahadat_2021}  & & 19.4M &  5.88G & 4.08ms & 56.10\\
\textbf{Axial-Hypercomplex}  & & \textbf{9.3M} & \textbf{1.36G} & \textbf{2.97ms} & \textbf{60.06-59.87} \\\hline

ResNet \cite{he2016deep} &  & 83.2M & 18.2G & 3.77ms & 59.06\\
ResNet-with-QPHM \cite{shahadat_2021} &  & 82.6M & 4.57G & 3.66ms & 60.30\\
Quaternion \cite{gaudet2018deep} & & 21.4M & 7.87G & 4.14ms & 56.63 \\
Vectormap \cite{gaudet2021removing} & 50 & 28.3M & 7.87G & 4.34ms & 57.52 \\
QPHM \cite{shahadat_2021}  &  & 20.8M & 1.93G & 3.88ms & 59.42\\
VPHM \cite{shahadat_2021}  &  & 27.7M & 7.75G & 4.51ms & 58.96\\
\textbf{Axial-Hypercomplex} &  & \textbf{13.7M} & \textbf{1.75G} & \textbf{3.93ms} & \textbf{62.73-62.07} \\\hline
\end{tabular} 
\caption{Image classification performance on the Tiny ImageNet benchmarks for 26, 35, and 50-layer architectures. Here, QPHM, and VPHM
define the quaternion networks with PHM FC layer, and vectormap networks with PHM FC layer, respectively.}
\label{tab_resultImageNet}
\end{table*}

\subsection{Method}
We conducted our experiments by using five-dimensional PHM dense layer in the backend of the network, quaternion network at the beginning of the network, and axial-hypercomplex residual bottleneck block on CIFAR benchmark datasets \cite{krizhevsky2009learning}, Street View House Numbers (SVHN) \cite{netzer2011reading}, and Tiny ImageNet \cite{Le2015TinyIV} datasets.

The models we tested to compare with our proposed model, are: the standard DCNNs \cite{he2016deep}, 
the DQNNs \cite{gaudet2018deep}, 
the axial-ResNet with QPHM \cite{shahadat_2021}, QPHM \cite{shahadat_2021}, VPHM \cite{shahadat_2021}, and our proposed method. CIFAR-10 and CIFAR-100 datasets consist of 60,000 color images of size 32 × 32 pixels. These datasets fall into 10 and 100 distinct classes and are split into a training set with 50,000 images and a test set with 10,000 images. We perform standard data augmentation schemes for these datasets like \cite{he2016deep,gaudet2018deep,gaudet2021removing,shahadat_2021}. Both datasets were normalized using per-channel mean and standard deviation. We perform horizontal flips and take random crops from images padded by 4 pixels on each side to obtain a 40 × 40 pixel image, then a 32 × 32 crop is randomly extracted. 

SVHN contains about 600,000 digit images \cite{netzer2011reading}. For experiments on SVHN we don’t do any image preprocessing, except simple mean/std normalization. We use similar augmentation for the Tiny ImageNet dataset which contains 100,000 training images of 200 classes (500 for each class) downsized to 64×64 colored images. The test set contains 10,000 images \cite{Le2015TinyIV}. 

All baseline models were trained using the same components as the real-valued networks, the original quaternion network, the original vectormap network, the QPHM, and the VPHM networks using the same datasets. All models in Table \ref{tab_resultCifar} were trained using the same
hyperparameters and the same number of output channels. The 50-layer architectural details of the above-mentioned models are depicted in Table \ref{tab_archiTable50} for the CIFAR-100 dataset. Due to space limitation, the deep ResNets and VPHM network architectures are not depicted in the architecture Table \ref{tab_archiTable50}. 

In the stem layer, the $3 \times 3$ convolution network is used for deep ResNets \cite{he2016deep}, $3 \times 3$ quaternion network is used for the deep quaternion ResNets\cite{trabelsi2017deep,gaudet2018deep}, for the QPHM  \cite{shahadat_2021}, and axial-hypercomplex networks (our proposed method), and $3 \times 3$ vectormap network is used for the deep vectormap ResNets \cite{gaudet2021removing}, and the VPHM \cite{shahadat_2021} networks with stride 1 \& 120 output filters. We use parameterized hypercomplex multiplication (PHM) for the dense layer in the backend of deep ResNets, QPHM, VPHM, and our proposed axial-hypercomplex networks. In the bottleneck block, the number of output channels of bottleneck groups are 120, 240, 480, \& 960 for all networks. In this experiment, we analyze 26-layer, 35-layer, and 50-layer architectures with the bottleneck block multipliers ``[1, 2, 4, 1]'', ``[2, 3, 4, 2]'', and ``[3, 4, 6, 3]''. These are depicted in Table \ref{tab_archiTable50}. 

We ran all of the models using stochastic gradient decent optimizer. We used linearly warmed-up learning from zero to 0.1 for the first 10 epochs and then used cosine learning rate scheduling from epochs 11 to 150. All models were trained for 128 batch sizes.

\subsection{Results}
The overall results of all models (base models and our proposed networks) appear in Tables \ref{tab_resultCifar}, \ref{tab_resultSVHN}, and \ref{tab_resultImageNet}. The top half of Table \ref{tab_resultCifar} shows the results for the CIFAR10 dataset and the bottom half presents the results for the  CIFAR100. Both datasets have been tested by the 26, 35, and 50 layers architectures. These are the parameter count, FLOPS count (number of multiply-add operations), inference time or Latency (time required to process a single image), and the percentage accuracy of validation results for each model. We evaluate original ResNets \cite{he2016deep}, ResNet with QPHM \cite{shahadat_2021}, original quaternion networks \cite{gaudet2018deep}, original vectormap networks \cite{gaudet2021removing}, QPHM \cite{shahadat_2021}, and VPHM \cite{shahadat_2021} with the same configuration like our proposed axial-hypercomplex networks. Our proposed axial-hypercomplex networks perform better in validation accuracy with lower parameter count and FLOPS for CIFAR-10 and CIFAR-100 datasets than the baseline networks. More precisely, our proposed method takes almost 6 times, 1/3 times, 1/2 times, 1/3 times, and 1/2 times fewer parameters than the ResNets, quaternion networks, vectormap networks, QPHM, and VPHM respectively. Moreover, axial-hypercomplex networks achieved state-of-the-art results for these CIFAR benchmarks in hypercomplex space. 

The performances for SVHN and Tiny ImageNet datasets are shown in Tables \ref{tab_resultSVHN} and \ref{tab_resultImageNet} for all architectures. The axial-hypercomplex network's validation accuracies outperform the other base networks with fewer trainable parameters and FLOPS like CIFAR datasets. The result Tables \ref{tab_resultCifar}, \ref{tab_resultSVHN}, and \ref{tab_resultImageNet}  show our proposed model performance ranges of three runs. However, the latency of axial-hypercomplex networks is a little bit higher in some cases than the quaternion-based networks. This may be due to the use of vectormap networks along with quaternion networks as the latency for vectormap networks is higher.

\section{Discussion and Conclusions}
This paper proposes axial-hypercomplex convolutions to reduce the cost of 2D convolutional operations and shows the effectiveness of image classification tasks. We also applied 4D PHM in the network's backend. 
On CIFAR benchmarks, our proposed Axial-hypercomplex network, formed by stacking axial-vectormap convolution (three-dimensional) in the quaternion bottleneck blocks, achieved state-of-the-art results among hypercomplex networks.

Our main conclusion is that using quaternion convolutions as the frontend stem layer, four/five-dimensional PHM-based densely connected backend layer, and axial-hypercomplex bottleneck block improves classification performance on the CIFAR benchmarks, SVHN, and Tiny ImageNet datasets in comparison to the other models we tested. Our proposed method factorizes a channel-wise 2D convolution (hypercomplex convolution which works along the channels) to a column convolution and a row convolution. Extensive experiments show that this leads to systematic improvement with far fewer trainable parameters on image classification. This proposed method can save 33\%, and 50\% trainable parameters compared to original quaternion and vectormap networks and QPHM and VPHM networks, respectively.

Although our proposed axial-hypercomplex design reduced parameter counts and FLOPS, it exhibited higher latency than real-valued and hypercomplex-valued convolutional networks. This is because the model performs convolution twice (height-axis and width-axis) and it takes transition time from 2D convolution to two consecutive 1D convolutions. As we replaced spatial quaternion (four-dimensional hypercomplex network) 2D convolution using two axial vectormap (three-dimensional hypercomplex network) 1D convolutions, the number of output channels are restricted to 120 or a multiple of 120 which are divisible by three and four. Our investigation concludes that the performance comparison between the hypercomplex networks and our proposed axial-hypercomplex networks shows that the axial-hypercomplex convolution provides better validation performance with fewer trainable parameters and FLOPS for image classification tasks. 

Further work may be directed toward the architecture of the axial quaternion network and axial vectormap network. Moreover, other datasets will be tested to check whether these proposed architectures can perform in a similar manner or not. Finally, axial-quaternion and axial-vectormap convolutional methods will 
help to remove the number of output channels constrained as it will divisible by four for axial-quaternion networks and three for axial-vectormap networks.

{\small
\bibliographystyle{ieee_fullname}
\bibliography{egbib}
}

\end{document}